\documentclass[letterpaper]{article} 
\usepackage{aaai2026}  
\usepackage{times}  
\usepackage{helvet}  
\usepackage{courier}  
\usepackage[hyphens]{url}  
\usepackage{graphicx} 
\usepackage{natbib}  
\usepackage{caption} 
\usepackage{subcaption} 
\usepackage{algorithm}
\usepackage{array}
\usepackage[utf8]{inputenc}
\usepackage{amsmath}
\usepackage{algorithm}
\usepackage{amsfonts}
\usepackage{amssymb}
\usepackage{caption}
\usepackage{algpseudocode}
\usepackage{tabularx}
\usepackage{ragged2e}

\usepackage{newfloat}
\usepackage{listings}
\usepackage{booktabs}
\usepackage{multirow}

\usepackage{newfloat}
\usepackage{listings}
\DeclareCaptionStyle{ruled}{labelfont=normalfont,labelsep=colon,strut=off} 
\floatstyle{ruled}
\newfloat{listing}{tb}{lst}{}
\floatname{listing}{Listing}
\title{SEVADE: Self-Evolving Multi-Agent Analysis with Decoupled Evaluation for Hallucination-Resistant Sarcasm Detection}
\author{
    Ziqi Liu\textsuperscript{\rm 1}\equalcontrib,
    Ziyang Zhou\textsuperscript{\rm 1}\equalcontrib,
    Yilin Li\textsuperscript{\rm 1},
    Mingxuan Hu\textsuperscript{\rm 1},
    Yushan Pan\textsuperscript{\rm 1},
    Zhijie Xu\textsuperscript{\rm 1},
    Yangbin Chen\textsuperscript{\rm 1}\thanks{Corresponding author.}
}
\affiliations{
    \textsuperscript{\rm 1}School of Advanced Technology, Xi'an Jiaotong-Liverpool University\\
    \{Ziqi.Liu, Ziyang.Zhou, Mingxuan.Hu\}22@student.xjtlu.edu.cn, Yilin.Li2202@student.xjtlu.edu.cn,\\ \{Yushan.Pan, Zhijie.Xu, Yangbin.Chen\}@xjtlu.edu.cn
}

\begin{document}

\maketitle

\begin{abstract}
Sarcasm detection is a crucial yet challenging Natural Language Processing task. Existing Large Language Model methods are often limited by single-perspective analysis, static reasoning pathways, and a susceptibility to hallucination when processing complex ironic rhetoric, which impacts their accuracy and reliability. To address these challenges, we propose \textbf{SEVADE}, a novel \textbf{S}elf-\textbf{Ev}olving multi-agent \textbf{A}nalysis framework with \textbf{D}ecoupled \textbf{E}valuation for hallucination-resistant sarcasm detection. The core of our framework is a Dynamic Agentive Reasoning Engine (DARE), which utilizes a team of specialized agents grounded in linguistic theory to perform a multifaceted deconstruction of the text and generate a structured reasoning chain. Subsequently, a separate lightweight rationale adjudicator (RA) performs the final classification based solely on this reasoning chain. This decoupled architecture is designed to mitigate the risk of hallucination by separating complex reasoning from the final judgment. Extensive experiments on four benchmark datasets demonstrate that our framework achieves state-of-the-art performance, with average improvements of \textbf{7.01\%} in Accuracy and \textbf{6.55\%} in Macro-F1 score.
\end{abstract}

\begin{links}
     \link{Code}{https://github.com/sunbus100/SEVADE}
\end{links}

\section{Introduction}

Sarcasm is defined as a rhetorical device or mode of expression, where the intended meaning of a statement is often the opposite of its literal meaning, frequently used to achieve a humorous or sarcastic effect \cite{booth1974rhetoric}.
Sarcasm detection aims to identify ironic expressions in text, which is crucial for improving applications such as sentiment analysis, content moderation, and public opinion mining\cite{joshi2017automatic,yi2025irony}.
However, sarcasm detection is a challenging Natural Language Processing (NLP) task because of its intrinsic dependence on contextual cues, the semantic incongruity between literal and intended meanings, and its multi-faceted nature which manifests through diverse linguistic phenomena like pragmatic contrast and emotional inversion\cite{grice1975logic,wilson1992verbal}.

With the advances of Computer Science and Linguistics, the pursuit of effective sarcasm detection has seen significant evolution.
Deep learning models used architectures such as CNNs, LSTMs, and GNNs to learn complex textual representations for training \cite{joshi2017automatic,liu2025caf}.
In particular, Transformer-based models, which used self-attention mechanisms and contextual embeddings, achieved remarkable performance in sarcasm detection by effectively capturing nuanced linguistic patterns and contextual cues within text \cite{gonzalez2020transformer}.
However, these paradigms struggled to grasp the implicit intent driving the intricate ironic rhetoric.

\begin{figure}
    \centering
    \includegraphics[width=1.0\linewidth]{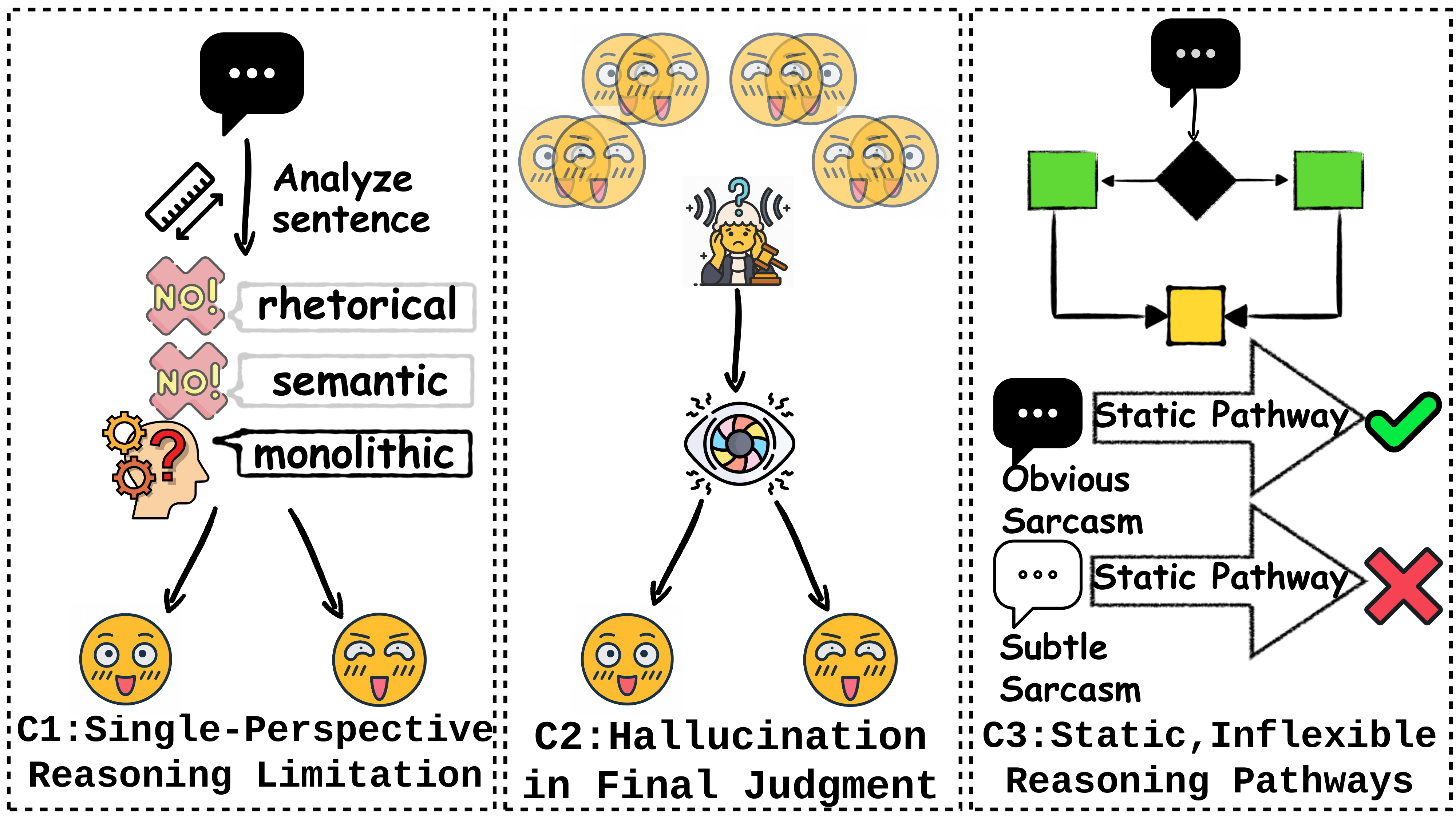}
    \caption{Key limitations of prior LLM-based sarcasm detection: single-perspective analysis, hallucination risk in final judgment, and static, inflexible reasoning pathways.}
    \label{problem_state}
\end{figure}

Recently, the emergence of Large Language Models (LLMs) has opened a promising new frontier in sarcasm detection, through both prompt engineering and instruction tuning \cite{liu2025caf,epron2024approach}. 
But there still remains three critical challenges which limit LLMs' efficacy and reliability in this nuanced domain:

\noindent
\textbf{C1: Single-Perspective Reasoning Limitation.} Standard LLM-based methods function as monolithic predictors, lacking the capacity to systematically deconstruct and analyze complex sarcasm from multiple linguistic dimensions\cite{guo2024large}.

\noindent
\textbf{C2: Hallucination Risk in Final Judgment.} LLMs are susceptible to hallucination, leading to unreliable or unfaithful judgments when synthesizing diverse and conflicting analytical signals into a single conclusion\cite{paul2024making}.

\noindent
\textbf{C3: Static and Inflexible Reasoning Pathways.} Current LLM-based sarcasm detection models often rely on fixed prompts or architectures, which limits their ability to dynamically adapt their analytical strategies to the specific complexities of the input\cite{zhang2024sarcasmbench,yao2025sarcasm}.

To address these limitations, we introduce a novel framework featuring a decoupled, multi-agent reasoning architecture. The process begins with our \textbf{D}ynamic
\textbf{A}gentive \textbf{R}easoning \textbf{E}ngine (\textbf{DARE}), which coordinates six specialized Core Analysis Agents. Each of the six specialized agents is grounded in a distinct linguistic perspective, selected based on linguistic theory and empirical studies~\cite{attardo2001humor,reyes2013multidimensional}, to capture the multi-dimensional cues crucial for robust sarcasm detection. The Controller Agent adaptively manages these agents through iterative refinement and targeted team expansion, allowing the reasoning depth and focus to dynamically adjust in response to the complexity of each input. Rather than producing a single prediction, DARE generates a structured reasoning chain, providing a transparent account of the underlying analytical process. This chain is subsequently evaluated by a lightweight Rationale Adjudicator, which delivers the final sarcasm judgment. By explicitly decoupling reasoning from decision-making, our framework mitigates hallucination risk and advances interpretability, surpassing conventional single-model paradigms.

The main contributions of this work are summarized as follows:
\begin{itemize}
    \item Propose a decoupled dynamic multi-agent sarcasm detection framework with adaptive, multi-perspective reasoning which emulates sophisticated human cognition.
    
    \item Enhance interpretability and mitigate hallucination via step-wise reasoning chains and separated adjudication.
    
    \item Achieve new SOTA on four sarcasm benchmarks, with superior robustness and generalization.
\end{itemize}

\section{Related Work}

\subsection{Multi-Agent Cooperation}
Multi-Agent Systems are a powerful paradigm for solving complex problems through collaborative interactions. Existing research has explored various cooperation patterns, such as debates, which enhance factual accuracy~\cite{du2023improving}, and structured dialogues for intricate problem-solving~\cite{wu2023mathchat}. Prominent frameworks like CAMEL employ role-playing to simulate nuanced behaviors~\cite{li2023camel}, while AutoGen offers high versatility through customizable conversation patterns~\cite{wu2023autogen}. While effective, current agent frameworks suffer from poor adaptability due to their fixed composition. The new research focus is on developing flexible agent systems capable of dynamic reasoning and self-adaptation.

\subsection{Self-Evolving Agents}
To overcome the limitations of static agent architectures, research has explored self-evolving systems. Initial approaches focused on dynamically adapting workflow or network topologies based on execution feedback~\cite{niu2025flow, yang2025agentnet}. More advanced paradigms involve agents that can rewrite their own source code for enhancement, as demonstrated in systems like ADAS~\cite{hu2024automated} and the Darwin-Gödel Machine~\cite{zhang2025darwin}, which leverage meta agents and empirical validation to drive open ended evolution. The core advantages of these self-evolving architectures such as adaptability are particularly valuable for pragmatic, context dependent language understanding tasks like sarcasm and sarcasm detection, where flexible and situation-aware analysis is crucial.

\subsection{Sarcasm Detection}
Sarcasm detection is a challenging NLP task due to its inherent contextual and contradictory nature. Early machine learning approaches relied on hand-crafted features, such as lexical cues~\cite{davidov2010semi} and sentiment lexicons~\cite{reyes2013multidimensional}, but struggled to capture the nuanced semantics of sarcasm. Subsequently, deep learning models leveraged word embeddings~\cite{mikolov2013efficient, pennington2014glove} and advanced architectures like CNNs~\cite{poria2016deeper}, LSTMs~\cite{zhang2016tweet}, and GNNs~\cite{liang2022multi} to learn hierarchical, sequential, and structural text features, yet often failed to capture the implicit intent behind complex ironic rhetoric. More recently, LLMs have been employed for sarcasm detection through prompting and zero-shot learning~\cite{yao2025sarcasm}. While promising, these methods typically function as single-model predictors, limiting their capacity for the multi-perspective reasoning necessary to fully comprehend intricate ironic expressions. Recent work has also explored multi-modal sarcasm detection by combining visual and textual cues~\cite{ramakrishnan2025ironic,zhang2025commander}. In parallel, CAF-I~\cite{liu2025caf} investigates collaborative multi-agent LLM frameworks for text-based sarcasm detection using a fixed agent committee. However, achieving adaptive reasoning in single-modality sarcasm detection remains a challenging problem. Therefore, we propose SEVADE, a dynamic and self-evolving multi-agent framework for robust and interpretable sarcasm detection.

\section{Methodology}
\begin{figure*}
    \centering
    \includegraphics[width=1.0 \linewidth]{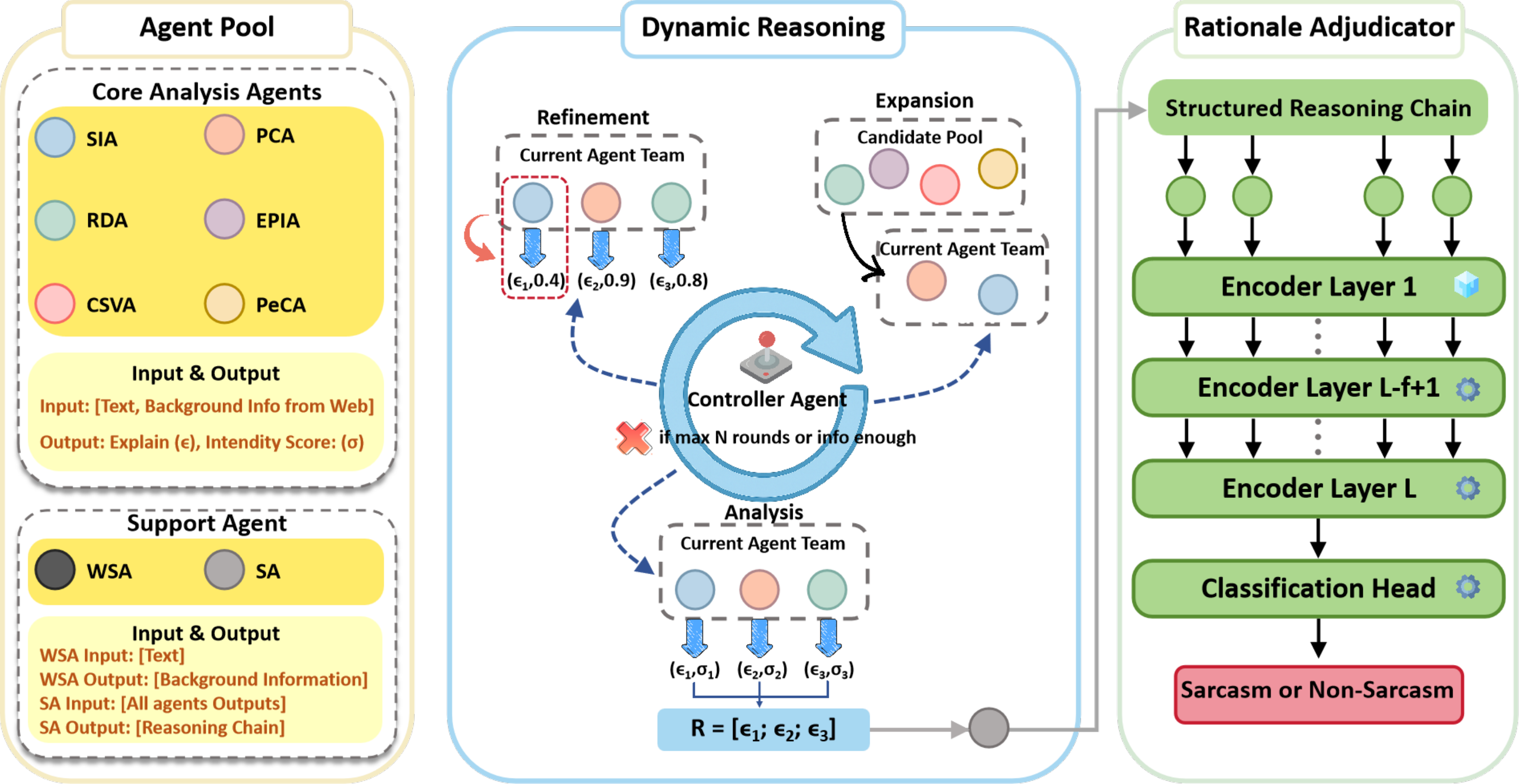}
    \caption{The overall framework of SEVADE. On the left, a pool of Core Analysis Agents, each with a distinct analytical perspective, and Support Agents with auxiliary functions are maintained. In the middle, the Controller Agent dynamically selects, refines, and expands the agent team to generate a structured reasoning chain. On the right, the Rationale Adjudicator produces the final sarcasm prediction based solely on the reasoning chain, decoupling reasoning from judgment.
}
    \label{structure}
\end{figure*}

\textbf{Problem Definition.}
Let the dataset be denoted as $\mathcal{D} = \{(c_i, y_i)\}_{i=1}^N$, where each instance consists of an input text $c_i$ and its corresponding ground truth label $y_i \in \{0, 1\}$, with $1$ indicating sarcasm. The primary objective of sarcasm detection is to learn a predictive function $\mathcal{G}$ that maps an input text $c$ to its corresponding label $y$, such that $\mathcal{G}: c \mapsto y$.

\textbf{Overall Framework.}
First, we design \textbf{DARE}, an iterative mechanism that generates a structured reasoning chain. The process unfolds as follows: (1)~\textbf{Instantiation.} A Controller Agent assembles an initial team of agents with analytical perspectives relevant to the input text. (2)~\textbf{Targeted Refinement.} In each cycle, the Controller identifies the single most ambivalent agent and prompts it to refine its analysis based on peer conclusions. (3)~\textbf{Adaptive Expansion.} If the reasoning stagnates, the Controller recruits a new agent from an inactive pool to introduce a complementary viewpoint. (4)~\textbf{Summarization.} Upon conclusion, a designated agent synthesizes the collective findings into a final reasoning chain $R$. 
Second, we introduce the Rationale Adjudicator, a lightweight classifier. It receives the reasoning chain $R$ as its sole input to perform the final classification, ensuring a robust prediction grounded in the preceding logical analysis. An overview of our proposed framework is shown in Figure~\ref{structure}.

\subsection{The Dynamic Agentive Reasoning Engine}
\subsubsection{Controller Agent and the Reasoning Process}
The reasoning process is governed by a central Controller Agent, which presides over the operational workflow of the DARE module to transform an input text into a final reasoning chain $R$. A foundational concept to this process is the structured output of each analytical agent, $a_i$. This output is a tuple, $O_i = (\sigma_i, \epsilon_i)$, comprising a scalar sarcasm intensity score, $\sigma_i \in [0, 1]$, representing perceived magnitude of the sarcastic signal, and a corresponding textual explanation, $\epsilon_i$.

The process is initiated with \textbf{instantiation}, where the Controller Agent adaptively selects the initial agent team, $\mathcal{A}_{\text{active}}^{(0)}$, from the full pool $\mathcal{A}_{\text{pool}}$, by prompting a language model to identify the analytical roles most relevant to the input text.

Following instantiation, the system enters an iterative cycle of \textbf{targeted refinement}. At each iteration $t$, the Controller identifies the single most ambivalent agent, $a_{\text{amb}}^{(t)}$, whose sarcasm intensity score exhibits maximum uncertainty.
\begin{equation}
a_{\text{amb}}^{(t)} = \underset{a_i \in \mathcal{A}_{\text{active}}^{(t)}}{\mathrm{argmin}} \left| \sigma_i^{(t)} - 0.5 \right|
\end{equation}
This designated agent is subsequently tasked with a refinement of its analysis. This refinement is executed via the function $\mathcal{F}_{\text{refine}}$, which operates upon the agent's prior output, $O_{\text{amb}}^{(t)}$, and is conditioned on the set of all peer outputs, yielding an updated analysis $O_{\text{amb}}^{(t+1)}$.
\begin{equation}
O_{\text{amb}}^{(t+1)} = \mathcal{F}_{\text{refine}}(O_{\text{amb}}^{(t)}, \{O_j^{(t)} \mid a_j \in \mathcal{A}_{\text{active}}^{(t)} \setminus \{a_{\text{amb}}^{(t)}\}\})
\end{equation}

Concurrently, the Controller performs a meta-cognitive check to determine if \textbf{adaptive expansion} is warranted. This check prompts the Controller to first render a judgment on whether the collective analysis is deemed \textit{incomplete, contradictory, or logically stuck}, and if affirmative, to then execute a meta-reasoning task to select the agent, $a_{\text{new}}$, from the inactive pool $\mathcal{A}_{\text{inactive}}^{(t)}$, that can most effectively address the identified analytical gaps. The active team is subsequently expanded:
\begin{equation}
\mathcal{A}_{\text{active}}^{(t+1)} = \mathcal{A}_{\text{active}}^{(t)} \cup \{a_{\text{new}}\}
\end{equation}
This iterative loop of refinement and expansion persists until the Controller Agent either deems the collective analysis to have achieved sufficient consistency. or when the pool of inactive agents is exhausted. The entire workflow is summarized in Algorithm~\ref{alg:reasoning}.

\begin{algorithm}[!h]
\caption{Dynamic and Iterative Reasoning Process}
\label{alg:reasoning}
\begin{algorithmic}[1]
    \State \textbf{Input:} Text $T$
    \State \textbf{Output:} Reasoning Chain $R$
    \State $\mathcal{A}_{\text{pool}} \gets \text{Initialize all Core \& Support Agents}$
    \State $\mathcal{A}_{\text{active}} \gets \text{Controller.InitialAgents}(T, \mathcal{A}_{\text{pool}})$
    \Loop
        \State $a_{\text{amb}} \gets \text{Controller.FindAmbivalentAgent}(\mathcal{A}_{\text{active}})$
        \State \text{ $a_{\text{amb}}$.refine\_analysis(context = $\mathcal{A}_{\text{active}}$) }
        \If{$\text{not Controller.ExpansionNeeded}(\mathcal{A}_{\text{active}})$}
            \State \textbf{break}
        \EndIf
        \State $\mathcal{A}_{\text{inactive}} \gets \mathcal{A}_{\text{pool}} \setminus \mathcal{A}_{\text{active}}$
        \If{$\mathcal{A}_{\text{inactive}} \text{ is empty}$}
            \State \textbf{break}
        \EndIf
        \State $a_{\text{new}} \gets \text{Controller.ComplementAgent}(T, \mathcal{A}_{\text{inactive}})$
        \State $\mathcal{A}_{\text{active}}.\text{add}(a_{\text{new}})$
    \EndLoop
    \State $R \gets \text{SummarizationAgent.summarize}(\mathcal{A}_{\text{active}})$
    \State \textbf{return} $R$
\end{algorithmic}
\end{algorithm}

\subsubsection{Agent Roles}

The DARE module is populated by a diverse set of agents engineered for specialized functions, which are categorized as either Core Analysis Agents or Support Agents.

\textbf{Core Analysis Agents.}
This agent pool consists of six agents grounded in linguistic and rhetorical theory, ensuring a multi-faceted deconstruction of the text.
\begin{itemize}
    \item \textbf{Semantic Incongruity Agent (SIA):} Identifies and quantifies conflicts between the text's literal meaning and established world knowledge.
    \item \textbf{Pragmatic Contrast Agent (PCA):} Analyzes the discordance between an utterance's formulation and its pragmatic context.
    \item \textbf{Rhetorical Device Agent (RDA):} Detects key figures of speech indicative of sarcasm, such as hyperbole and understatement.
    \item \textbf{Emotion Polarity Inverter Agent (EPIA):} Measures the contradiction between the text's overtly expressed emotion and the sentiment that would be objectively inferred from the situation.
    \item \textbf{Common Sense Violation Agent (CSVA):} Evaluates if the text's content violates widely held principles of common sense.
    \item \textbf{Persona Conflict Agent (PeCA):} Examines and reports on inconsistencies between the speaker's projected persona and the content of their statement.
\end{itemize}

\textbf{Support Agents.}
This category includes agents that provide essential auxiliary functions to the reasoning process. The \textbf{Web Search Agent (WSA)} is selectively invoked before the main analysis stage, based on a decision by the Controller Agent. Given the input text, if external context is deemed necessary, the Controller uses an extraction function $\mathcal{F}{\text{extract}}$ to generate one or two concise keywords:
\begin{equation}
\mathcal{K} = \mathcal{F}_{\text{extract}}(T)
\end{equation}
These keywords are used by the WSA to query external sources and retrieve the top three most relevant background knowledge snippets. The retrieved evidence, together with the original text, is then provided to each analysis agent, supporting more informed and context aware reasoning. In contrast, the \textbf{Summarization Agent (SA)} executes the final step of the DARE workflow, synthesizing the structured output of all active agents into a single, coherent reasoning chain $R$.

\subsection{Rationale Adjudicator}
The Rationale Adjudicator (RA) is our framework's final component, designed to robustly adjudicate the reasoning chain generated by the DARE module. To mitigate the hallucination risks inherent in a large model's synthesis of diverse analyses, we employ a lightweight, fine-tuned model for this task, ensuring reliable and specialized final judgment. A critical design choice is that the Rationale Adjudicator's sole input is the reasoning chain, $R$, generated by the DARE module. This forces the model to base its judgment exclusively on the logical coherence and semantic patterns of the provided rationale.

We implement the Rationale Adjudicator with BERT~\cite{devlin2019bert}, fine-tuning only the parameters of the last $f$ layers while freezing all preceding layers. In all experiments, we set $f=2$: 
\begin{equation}
\mathcal{A}_{\mathrm{RA}}(\mathbf{x}) = \mathcal{L}_L \circ \mathcal{L}_{L-1} \circ \cdots \circ \mathcal{L}_{L-f+1} \circ \mathcal{L}^{\text{fixed}}_{L-f} \circ \cdots \circ \mathcal{L}^{\text{fixed}}_{1}(\mathbf{x})
\end{equation}
This strategy allows the model to adapt its high-level representations for the specific task of interpreting reasoning chains. The entire model is then optimized by minimizing the Binary Cross-Entropy loss, $\mathcal{L}_{\text{BCE}}$, between its prediction $\hat{y}$ and the ground-truth label $y$.
\begin{equation}
\mathcal{L}_{\text{BCE}} = -\frac{1}{N} \sum_{i=1}^{N} \left[ y_i \cdot \log(\hat{y}_i) + (1 - y_i) \cdot \log(1 - \hat{y}_i) \right]
\end{equation}

\section{Experiments}
\subsection{Experiment Setup}
An overview of the experimental setup is outlined below.

\noindent
\subsubsection{Datasets}
We conduct evaluations on four sarcasm detection benchmarks. Includes \textbf{IAC-V1} \cite{lukin2013really}, \textbf{IAC-V2} \cite{oraby2017creating}, \textbf{MuSTARD} and \textbf{SemEval-2018 Task 3} \cite{van2018semeval}. Statistics of datasets are summarized in Table~\ref{table:dataset_stats}. 

\begin{table}[!htp]
    \centering
    \renewcommand{\arraystretch}{0.8}
    \small
    \begin{tabular}{lcccc}
        \toprule
        \textbf{Dataset} & \textbf{Train} & \textbf{Val} & \textbf{Test} & \textbf{Avg. Len}  \\
        \midrule
        IAC-V1  & 1595 & 80 & 320 & 68 \\
        IAC-V2 & 5216 & 262 & 1042 & 43 \\
        SemEval-2018  & 3634 & 200 & 784 & 14 \\
        MuSTARD & 552 & -- & 138 & 14 \\
        \bottomrule
    \end{tabular}
    \caption{Overview of the benchmark datasets used for evaluating sarcasm detection.}
    \label{table:dataset_stats}
\end{table}

\noindent
\subsubsection{Baselines}
We conduct a comparative analysis against a diverse suite of baselines organized into three main categories. \textbf{LLM-Based}: We include approaches from SarcasmCue~\cite{yao2025sarcasm}, such as GPT-4o zero-shot and three prompting strategies, Chain of Contradiction, Graph of Cues, and Bagging of Cues, with performance reported from their original results. We also add GPT-5 as a stronger reasoning baseline for completeness. \textbf{Fine-tuned PLMs}: We evaluate BERT-base~\cite{devlin2019bert} and RoBERTa-base~\cite{liu2019roberta}, both fine-tuned on each target dataset. \textbf{Deep Learning Methods}: We compare with representative methods including MIARN~\cite{tay2018reasoning}, SAWS~\cite{pan2020modeling}, and DC-Net~\cite{liu2021dual}, with results taken from previous studies~\cite{qiu2024detecting,zhang2024sarcasmbench,tay2018reasoning,xue2024breakthrough,pan2020modeling}.

\noindent
\subsubsection{Metrics and Implementation Details}
Our framework employs the \textbf{GPT-4o} model as backbone for all agents, accessed via the official OpenAI API with a temperature of 0 to ensure reproducibility. Following standard practices \cite{yao2025sarcasm}, performance is evaluated using \textbf{Accuracy} and primarily the \textbf{Macro-F1 score}.

\subsection{Main Result}
\begin{table*}[!htp]
    \centering
    \fontsize{9}{11}\selectfont
    \renewcommand{\arraystretch}{0.75} 
    \begin{tabular*}{\textwidth}{@{\extracolsep{\fill}}>{\raggedright\arraybackslash}p{1.8cm} cc cc cc cc cc@{}}
        \toprule
         & \multicolumn{2}{c}{\textbf{IAC-V1}} & \multicolumn{2}{c}{\textbf{IAC-V2}} & \multicolumn{2}{c}{\textbf{MuSTARD}} & \multicolumn{2}{c}{\textbf{SemEval-2018}} & \multicolumn{2}{c}{\textbf{Avg.}} \\
        \cmidrule(lr){2-3} \cmidrule(lr){4-5} \cmidrule(lr){6-7} \cmidrule(lr){8-9} \cmidrule(lr){10-11}
        \textbf{Baseline} & \textbf{Acc.} & \textbf{M-F1} & \textbf{Acc.} & \textbf{M-F1} & \textbf{Acc.} & \textbf{M-F1} & \textbf{Acc.} & \textbf{M-F1} & \textbf{Acc.} & \textbf{M-F1}\\
        \midrule
        MIARN & 0.6321 & 0.6318 & 0.7275 & 0.7275 & 0.6460 & 0.6390 & 0.6850 & 0.6780 & 0.6726 & 0.6691 \\
        SAWS  & 0.6613 & 0.6560 & 0.7620 & 0.7620 & 0.6971 & 0.7095 & 0.6990 & 0.6890 & 0.7048 & 0.7041\\
        DC-Net & 0.6650 & 0.6640 & \underline{0.7800} & \underline{0.7790} & \underline{0.7128} & \underline{0.7143} & \underline{0.7630} & \underline{0.7670} & \underline{0.7302} & \underline{0.7311} \\
        \midrule
        BERT  & 0.6530 & 0.6520 & 0.7640 & 0.7620 & 0.6430 & 0.6430 & 0.6990 & 0.6840 & 0.6897 & 0.6852 \\
        RoBERTa  & 0.7010 & 0.6990 & 0.7660 & 0.7670 & 0.6610 & 0.6600 & 0.7020 & 0.6910 & 0.7075 & 0.7042\\
        \midrule
        GPT-4o  & 0.7063 & 0.7005 & 0.7303 & 0.7199  & 0.6724 & 0.6579 & 0.6403 & 0.6317 & 0.6873 & 0.6775\\
        GPT-4o+CoC  & \underline{0.7219} & \underline{0.7152} & 0.7336 & 0.7231 & 0.6942 & 0.6848 & 0.7079 & 0.7060 & 0.7144& 0.7073 \\
        GPT-4o+GoC & 0.6500 & 0.6291 & 0.6497 & 0.6130 & 0.7069 & 0.6991 & 0.7403 & 0.7402  &  0.6867& 0.6704\\
        GPT-4o+BoC & 0.6875 & 0.6736 & 0.7135 & 0.6939 & 0.6942 & 0.6845 & 0.6212 & 0.6185 & 0.6791 & 0.6676\\
        GPT-5  & 0.7031 & 0.7084 & 0.7493 & 0.7442 & 0.7076 & 0.6994 & 0.7257 & 0.7183 & 0.7214 & 0.7178 \\
        \midrule
        \textbf{Ours} & \textbf{0.7304} & \textbf{0.7298} & \textbf{0.7832} & \textbf{0.7830} & \textbf{0.7681} & \textbf{0.7638} & \textbf{0.8439} & \textbf{0.8394} & \textbf{0.7814}& \textbf{0.7790} \\
        Improv. & 1.17\%$\uparrow$ & 2.04\%$\uparrow$ & 0.51\%$\uparrow$ & 0.41\%$\uparrow$ & 7.75\%$\uparrow$ & 6.92\%$\uparrow$ & 10.61\%$\uparrow$ & 9.43\%$\uparrow$ & 7.01\%$\uparrow$& 6.55\%$\uparrow$\\
        \bottomrule
    \end{tabular*}
        \caption{Overall performance comparison across four benchmark datasets. All LLM strategies are zero-shot. \textbf{Acc.} denotes Accuracy and \textbf{Ma-F1} signifies Macro-F1. Best results are in \textbf{bold}, second-best are \underline{underlined}. Scores are reported as decimals.}
        \label{tab:main_results}
\end{table*}

Table \ref{tab:main_results} presents a comprehensive comparison of our framework against various baselines. From the results, we have the following observations.

First, our framework establishes a new state-of-the-art across all datasets, achieving an average Accuracy of 78.14 and Macro-F1 of 77.90, improving by 7.01\% and 6.55\% over the strongest baseline, DC-Net. This performance stems from our synergistic two-stage architecture, the DARE module generates structured rationales through multi-agent analysis, and the lightweight Rationale Adjudicator classifies solely based on them. This design decouples reasoning from judgment, ensuring the decision is grounded in explicit logic and mitigating hallucination risks. Even against strong reasoning model like GPT-5, SEVADE remains clearly superior, confirming its design advantage.

Second, our findings reveal a core weakness in generalist LLMs, they lack the specific task adaptation required for nuanced sarcasm detection, while fine-tuning is computationally prohibitive and impractical. Our framework is designed to resolve this exact trade-off through the specific mechanism of the DARE module. By structuring the reasoning process around a dynamic, multi-perspective analysis, DARE provides an effective form of task adaptation. This targeted approach explains its ability to robustly handle complex cases where monolithic LLM architectures falter.

Finally, SEVADE’s advantage is most evident on the complex MuSTARD and SemEval datasets, with accuracy gains of 7.75\% and 10.61\% over the strongest baseline. This suggests these benchmarks require reasoning that transcends surface-level pattern matching, often hinging on common sense or external world knowledge. Traditional models lack mechanisms for such explicit reasoning. Our framework addresses this gap directly; agents like the Common Sense Violation Agent and the on-demand Web Search Agent explicitly integrate external validation, enabling our model to resolve intricate sarcastic expressions that confound other approaches.

\subsection{Ablation Study}
\begin{table}[h!]
\centering
\fontsize{9}{11}\selectfont
\renewcommand{\arraystretch}{0.7} 
\begin{tabular*}{\columnwidth}{@{\extracolsep{\fill}}lcccc@{}}
\toprule
\textbf{Dataset} & \multicolumn{2}{c}{\textbf{IAC-V1}} & \multicolumn{2}{c}{\textbf{Semeval-2018}} \\
\cmidrule(r){2-3} \cmidrule(l){4-5}
\textbf{Variant}& \textbf{Acc.} & \textbf{M-F1} & \textbf{Acc.} & \textbf{M-F1} \\
\midrule
w/o SIA & 0.6708 & 0.6617 & 0.8091 & 0.8059 \\
w/o PCA & 0.6783 & 0.6747 & 0.7176 & 0.7166 \\
w/o RDA & 0.6891 & 0.6889 & 0.7734 & 0.7734 \\
w/o EPIA & 0.6825 & 0.6808 & 0.7361 & 0.7352 \\
w/o CSVA & 0.7047 & 0.7033 & 0.7328 & 0.7319 \\
w/o PeCA & 0.7038 & 0.7031 & 0.7349 & 0.7342 \\
\midrule
w/o Evolving & 0.6928 & 0.6976 & 0.7754 & 0.7754\\
w/o RA & 0.7147 & 0.7138 & 0.7558 & 0.7555 \\
\midrule
\textbf{Full Model} & \textbf{0.7304} & \textbf{0.7298} & \textbf{0.8439} & \textbf{0.8394} \\
\bottomrule
\end{tabular*}
\caption{Ablation study on key components of our framework.}
\label{tab:my_label}
\end{table}
We comprehensively validate our framework with the ablation study detailed in Table~\ref{tab:my_label}. We first assess the contribution of six core analysis agent by creating variants in which they are individually removed. Furthermore, we evaluate the core architecture through two mechanism-focused variants: \textbf{w/o Evolving}, which forces a static analysis by disabling iterative refinement and expansion; and \textbf{w/o RA}, which replaces our specialized Rationale Adjudicator with the base LLM to classify the final reasoning chain $R$.

The experiment results highlight the following conclusions:
\textbf{(1) Each Core Analysis Agent provides essential insights.} The removal of any single agent leads to a noticeable drop in performance across all datasets. This demonstrates that our chosen set of agents offers a comprehensive, non-redundant analytical foundation, and a multi-faceted approach is critical for robust sarcasm detection;
\textbf{(2) The impact of agents is dataset dependent.} For IAC-V1, performance is most affected by the absence of the Semantic Incongruity and Pragmatic Contrast agents. For Semeval-2018, the Pragmatic Contrast and Emotion Polarity Inverter agents are the most influential, highlighting the need for context and sentiment analysis for this dataset;
\textbf{(3) The dynamic evolution process is crucial.}
 The \textbf{w/o Evolving} variant, which uses a static analysis, shows a substantial performance drop on both datasets. This result confirms that our iterative process of refinement and expansion is superior to a static agent committee approach;
\textbf{(4) The superior performance of our full model over the w/o RA variant.} This finding validates our specialized Rationale Adjudicator, unlike the general-purpose base LLM, our lightweight classifier's constrained nature mitigates hallucination risks, ensuring a more reliable judgment that is faithfully grounded in the preceding analysis.

\subsection{Model Explainability}
\begin{figure}[!htp] 
    \centering 
    \begin{subfigure}[t]{0.48\columnwidth}
        \centering
        \includegraphics[width=\linewidth]{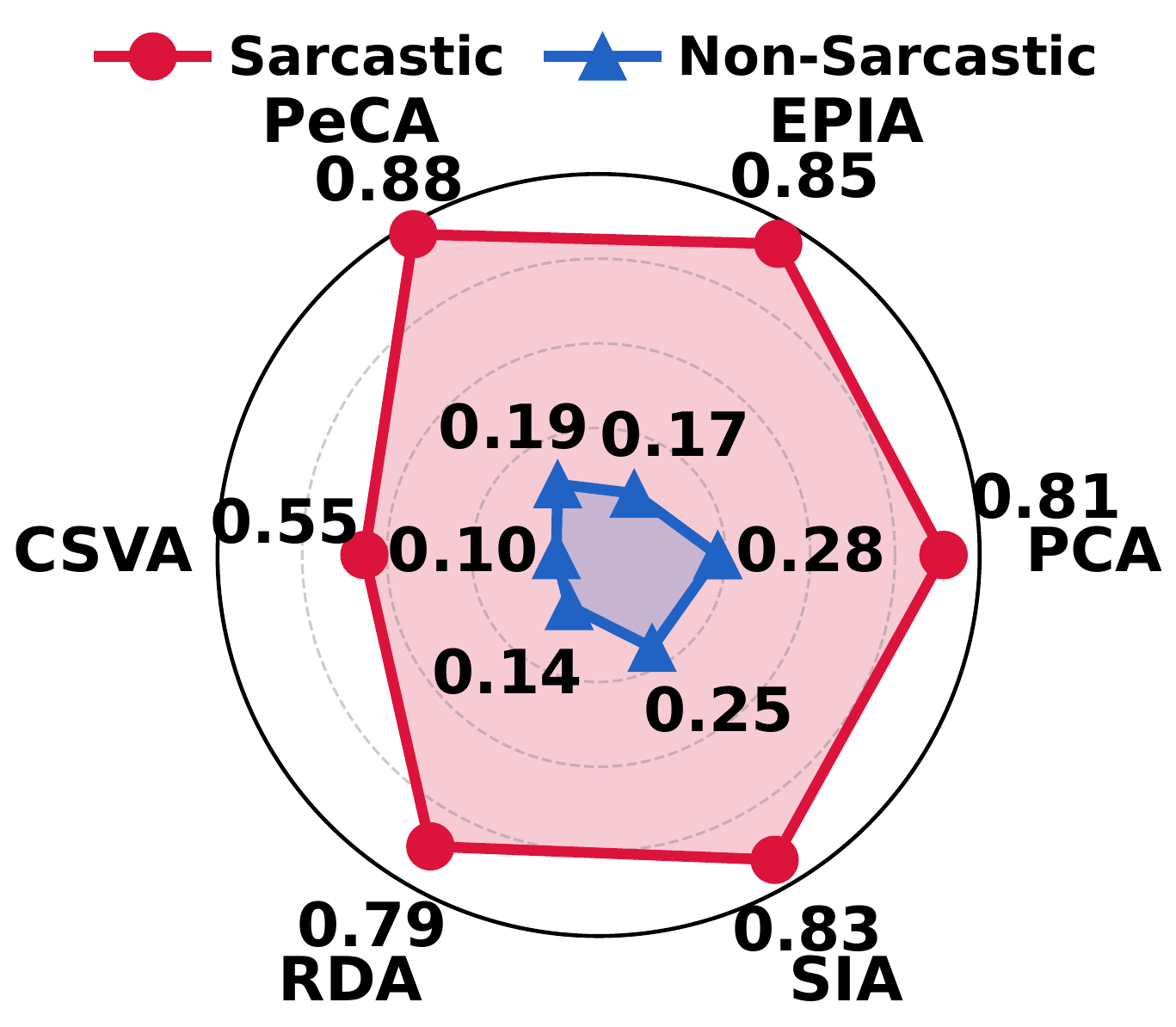}
        \caption{Mean intensity of Agents.}
        \label{fig:radar_contribution} 
    \end{subfigure}
    \hfill 
    \begin{subfigure}[t]{0.48\columnwidth}
        \centering
        \includegraphics[width=\linewidth]{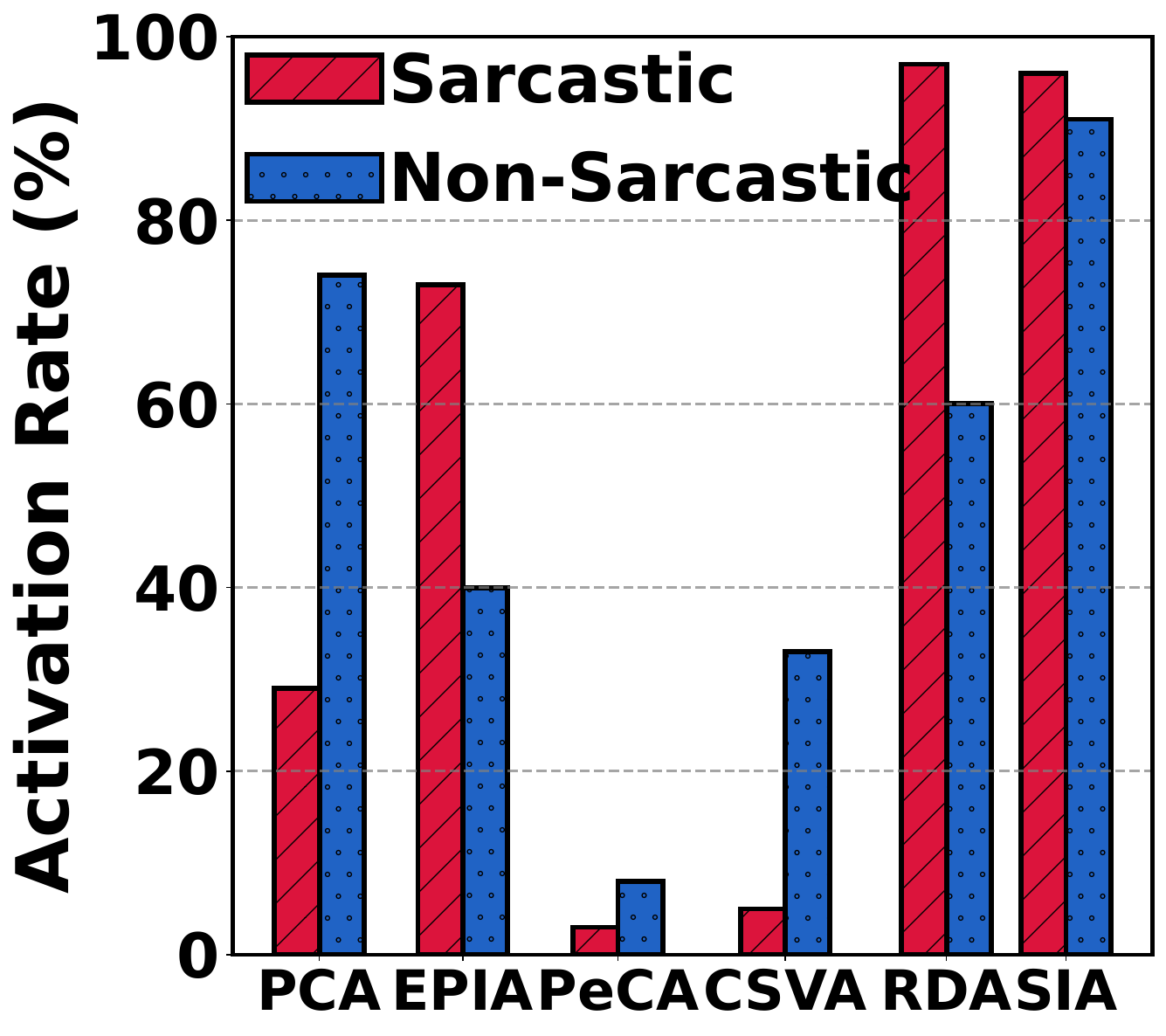}
        \caption{Activation Rate of Agents.} 
        \label{fig:bar_activation} 
    \end{subfigure}
    \caption{Visualization of agent dynamics in processing sarcastic and non-sarcastic samples. (a) shows the mean intensity score. (b) shows the activation rate, indicating agent participation frequency.}
    \label{fig:agent_dynamics}
\end{figure}
We also investigate the internal dynamics of the DARE framework to validate the adaptability and interpretability of its process by analyzing agent behavior (Figure \ref{fig:agent_dynamics}).

From the result we present two key findings: \textbf{(1) Agents are highly specialized.} While all agents show baseline sensitivity by assigning appropriate high/low intensity scores when dealing with sarcasm/non sarcasm sample \ref{fig:radar_contribution}, they exhibit profound functional specialization. The RDA and SIA agents dominate sarcastic detection with both high activation rates and the highest intensity scores , acting as primary incongruity detectors. Conversely, the PCA is more likely to be activated for non-sarcastic samples as a dedicated contextual congruity verifier; \textbf{(2) The model follows an evolutionary reasoning process driven by refinement and expansion.} This strategy is evidenced by the differential activation patterns. The model's reasoning process evolves by shifting its analytical focus based on initial evidence. For sarcastic samples containing strong signals, the framework converges efficiently without needing extensive contextual analysis from PCA. For non-sarcastic samples that lack such obvious signals, the framework adaptively invokes PCA to perform a deeper contextual non-sarcasm validation. In short, this evolution is co-driven by refinement on a single perspective for analytical depth, and expansion with new perspectives for conceptual breadth. In conclusion, DARE's evolutionary reasoning process culminates in an interpretable reasoning chain, which shows how the decision evolved.

\subsection{Model Generalizability}
\begin{figure}[!htp]
    \centering
    \includegraphics[width=1.0\linewidth]{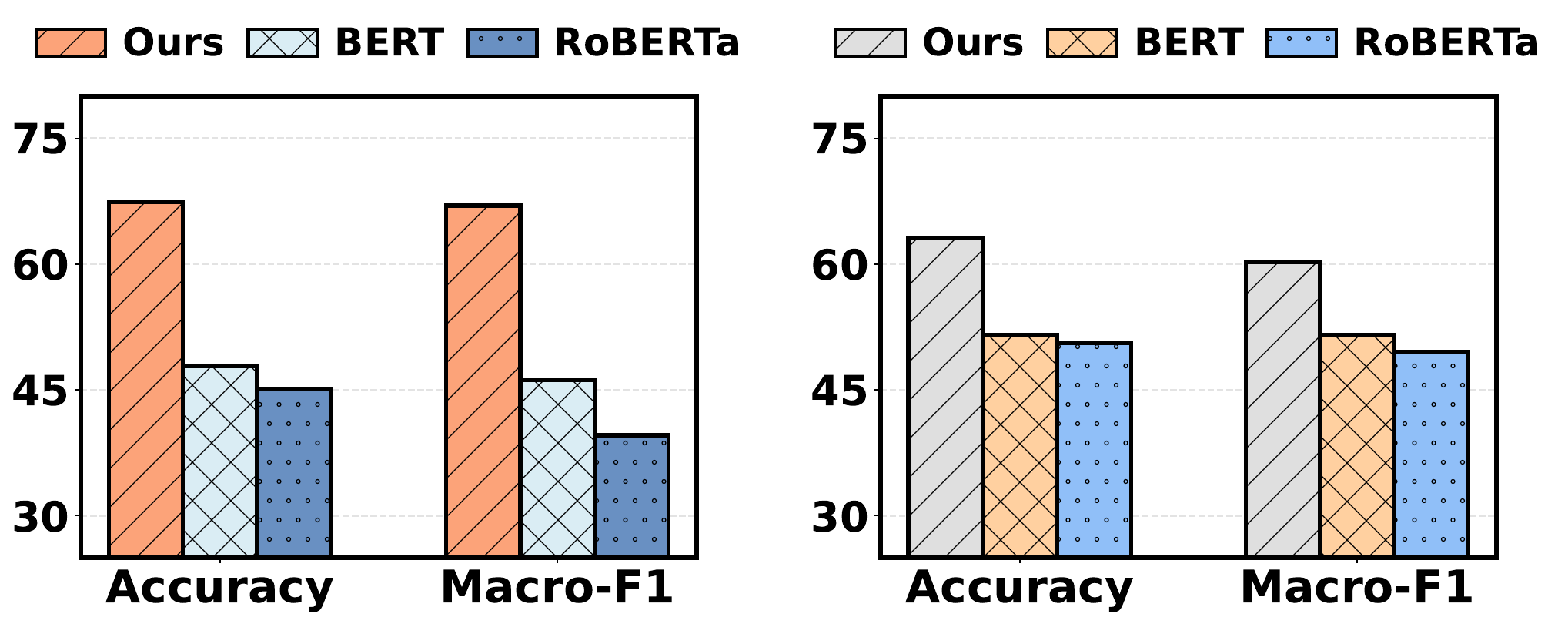}
    \caption{Cross-dataset generalization performance. The left chart shows model that train on the IAC-V1 and test on semeval. The right chart shows the reverse scenario. Our proposed model is compared against BERT and RoBERTa.}
    \label{fig:enter-label}
\end{figure}

\begin{table*}[!htbp]
\centering
\begin{tabularx}{\textwidth}{@{}l c >{\RaggedRight}X@{}}
\toprule
\multicolumn{3}{@{}l@{}}{
  \begin{minipage}[t]{\linewidth} 
    \setlength{\parindent}{0pt} 
    \textbf{Input Text: }“Do you mean someone who was wrong? I asked you before to give an example of where Newton found empirical evidence of God. You failed to do so. I suspect that you can't.”
  \end{minipage}
} \\
\multicolumn{3}{@{}l@{}}{\textbf{Ground Truth:} NOT SARCASTIC \qquad \textbf{Model Prediction:} SARCASTIC} \\
\midrule
\textbf{Agent} & \textbf{Intensity Score} & \textbf{Agent's Rationale} \\
\midrule

\textbf{SIA} & 
~0.8 & 
"... due to the accusatory tone ... and the \textbf{mismatch between the literal question and the speaker's clear skepticism}, which implies dismissal ..." \\
\addlinespace 

\textbf{PCA} & 
~0.2 & 
"The tone reads as critical but ... straightforward, with \textbf{no strikingly jarring mismatch between the situation (a debate ...) and the language style}. It seems literal rather than sarcastic." \\
\addlinespace

\textbf{RDA} & 
~0.8 & 
"... Pragmatic Contrast Agent highlights the critical tone ... Refining my explanation, \textbf{the rhetorical question 'Do you mean someone who was wrong?' mock the possibility of evidence} ..." \\

\addlinespace

\midrule

\multicolumn{3}{@{}l@{}}{
    \textbf{Summarization:} "The agents predominantly indicate sarcasm conveyed through skepticism and rhetorical dismissal \ldots "
} \\

\bottomrule
\end{tabularx}
\caption{Case study of a false positive sample on the IAC-V1 dataset.}
\label{tab:case_study_23_final}
\end{table*}

To assess our framework's real world applicability, we conducted a rigorous cross-dataset evaluation, training on one dataset and testing on the other. This setup evaluates the model's ability to generalize across different data distributions and linguistic styles against BERT and RoBERTa baselines. As detailed in Figure \ref{fig:enter-label}, our framework demonstrates superior generalization. For instance, when trained on IAC-V1 and tested on semeval, our model achieves an M-F1 score of 66.97, surpassing the RoBERTa by over 27\%. This significant performance advantage is consistent in the reverse setting as well, highlighting our model's robustness.

This superior robustness stems directly from our framework's design philosophy. Unlike monolithic models that overfit to superficial textual features, our approach first employs a set of specialized agents. These agents are explicitly designed based on formal linguistic and semantic principles, enabling them to extract robust, generalizable signals of sarcasm rather than dataset-specific artifacts. The pivotal Summarization Agent then synthesizes these high-quality, multi-perspective analyses into a structured reasoning chain $R$. By operating on this clean and logically coherent representation, the final Rationale Adjudicator makes its prediction based on universal sarcastic patterns, effectively overcoming the challenges of domain shift.

\subsection{Influences of LLM scales}
\begin{figure}[!htbp]
    \centering
    \includegraphics[width=1.0\linewidth]{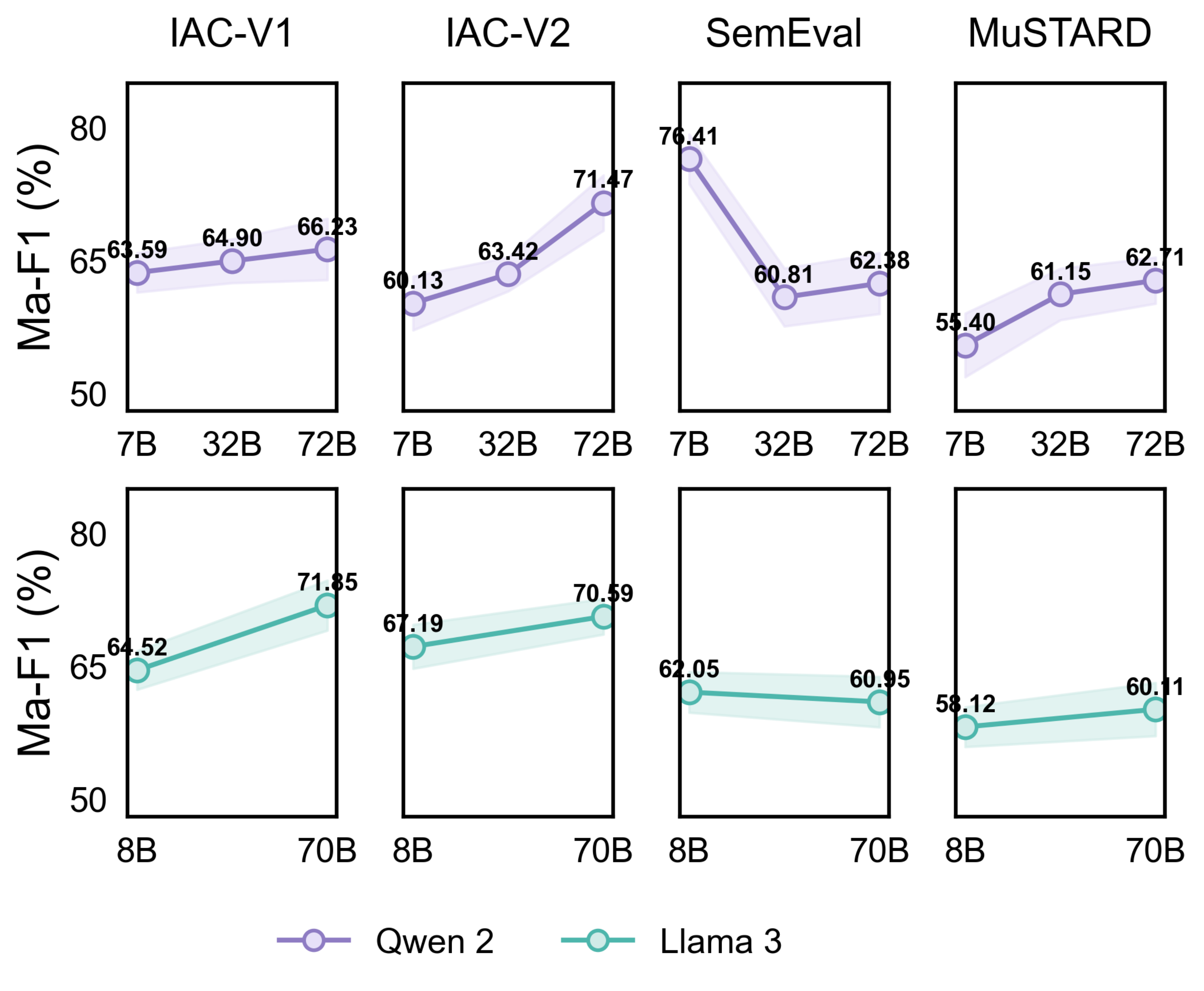}
    \caption{The effect of model scale on performance, using the Qwen 2 and Llama 3 series.}
    \label{model_scale}
\end{figure}
In our examination of model scale's influence, a clear pattern emerges across most datasets. As illustrated in Figure~\ref{model_scale}, the performance of our framework generally improves as the underlying LLM's parameter count increases. This aligns with the understanding that larger models possess more sophisticated reasoning capabilities, and enable DARE to produce more nuanced analyses.

Conversely, the results on the SemEval dataset present an important counter-trend, where smaller models achieve superior performance. This outcome stems from our two-stage architectural design. On datasets characterized by explicit cues like SemEval, smaller models generate concise and logically direct reasoning chains, which our rationale adjudicates reliably. Larger models, in contrast, exhibit a propensity for over-analysis, producing convoluted rationales laden with secondary or spurious details. Our adjudicator evaluates these less coherent chains as lower quality, leading to diminished performance. This phenomenon illustrates our framework's architectural strength, its efficacy is determined by the coherence of the generated reasoning rather than the raw parameter scale. This inherent robustness underscores the framework's adaptability and its utility in resource-constrained scenarios.

\subsection{Error Analysis}
Our error analysis indicates a predominant failure mode of misclassifying non-sarcastic texts as sarcastic, resulting in a higher false positive rate, a finding that is consistent with observations in similar advanced models\cite{yao2025sarcasm}. To investigate this common over-detection challenge, we deconstruct a representative case in Table \ref{tab:case_study_23_final}. 

From the case we can see that while SIA and RDA identified strong textual cues for sarcasm, such as \textit{mismatch between the literal question and the speaker's clear skepticism} and a \textit{rhetorical question} used to \textit{mock the possibility of evidence} the PCA provided crucial counterevidence that language is appropriate for debate circumstance. The actual misclassification, which results from this benign internal disagreement, highlights a core direction for future research, developing more advanced, context-sensitive perspective fusion mechanisms that go beyond simple cue identification or weighting, enabling models to flexibly and holistically integrate multiple linguistic perspectives. 

\section{Conclusion}
In this work, we proposed SEVADE, a novel multi-agent framework that overcomes key sarcasm detection challenges like single-perspective, static analysis and LLM hallucination. Its core, the DARE, uses specialized linguistic agents to perform a multi-faceted analysis, generating a structured reasoning chain. A separate, lightweight Rationale Adjudicator then makes the final judgment based solely on this rationale, a decoupled architecture that effectively mitigates hallucination. Our approach achieves new state-of-the-art performance across four benchmarks, demonstrating superior accuracy and generalization.

\section*{Acknowledgements}
This research is supporte by XJTLU RDF-24-02-008, National Natural Science Foundation of China (Grant No. 62502396), and Leadership Talent Program (Science and Education) of SIP, KJL2024104.
\bibliography{aaai2026}
\end{document}